\documentclass[runningheads]{llncs}
\usepackage[T1]{fontenc}
\usepackage{graphicx}
\usepackage{graphicx}
\usepackage{amsmath,amssymb}
\usepackage{booktabs}
\usepackage{url}
\usepackage{algorithm}
\usepackage{algorithmic}
\usepackage{pifont}
\usepackage{array}
\usepackage{multirow}

\begin{document}
%
% \title{FP4Quant: 4-Bit Ultra-Low Precision Communication for Model-Parallel LLM Training}
\title{FP4Quant: Efficient 4-Bit Communication Compression for Model-Parallel LLM Training}
\author{Man Liu\inst{1}\orcidID{0009-0006-7390-0220} \and
Xingchen Liu\inst{2}\orcidID{0009-0006-9410-5365} \and Hairui Zhao\inst{2}\orcidID{0009-0008-7081-5172} \and Guangming Tan\inst{2}\orcidID{0000-0002-6361-5948} \and
Dingwen Tao\inst{2}\orcidID{0000-0001-5422-4497}}

\authorrunning{M. Liu et al.}

\institute{
University of Chinese Academy of Sciences, Beijing, China
\and
Institute of Computing Technology, Chinese Academy of Sciences, Beijing, China
\email{liuman24@mails.ucas.ac.cn, liuxingchen232@mails.ucas.ac.cn, zhaohairui@ict.ac.cn, tgm@ict.ac.cn,  taodingwen@ict.ac.cn}
}

\titlerunning{FP4Quant}
% \titlerunning{FP4Quant: Efficient 4-Bit Communication Compression for LLM Training}

% \author{Man Liu\inst{1}\orcidID{0009-0006-7390-0220} \and
% Xingchen Liu\inst{2,3}\orcidID{0009-0006-9410-5365} \and
% Dingwen Tao\inst{3}\orcidID{0000-0001-5422-4497}}

% % % First names are abbreviated in the running head.
% % % If there are more than two authors, 'et al.' is used.
% % %
% \institute{University of Chinese Academy of Sciences, No.1 Yanqihu East Rd, Huairou District, Beijing, PR China, 101408 \and
% Institute of Computing Technology, Chinese Academy of Sciences, 6 Kexueyuan Nanlu, Zhongguancun, Haidian, Beijing, China, 100190

% \author{Author Name}
% \institute{Affiliation, City, Country\\ \email{author@example.com}}

\maketitle

\begin{abstract}

The rapid scaling of large language models (LLMs) has made communication overhead a critical bottleneck in distributed model-parallel training. Although existing low-precision compression methods can effectively reduce communication costs, further reducing the communication footprint to ultra-low-bit formats such as FP4 introduces new challenges, including limited numerical range, sub-byte storage overhead, and increased computational cost from fine-grained quantization. In this paper, we present \textbf{FP4Quant}, a micro-grouped FP4 compression framework for efficient model-parallel LLM training. FP4Quant integrates a micro-grouped dual-scale FP4 quantization mechanism to preserve local tensor distribution characteristics, a warp-cooperative FP4 packing scheme to enable efficient sub-byte data representation, and a hybrid two-level FWHT pipeline to mitigate quantization errors with GPU-aware execution optimization. 
Experiments on NVIDIA H100 GPUs with Qwen-7B demonstrate up to 1.56$\times$ end-to-end throughput improvement over BF16, revealing a trade-off between throughput and model quality.


\end{abstract}

\section{Introduction}

In recent years, large language models (LLMs) have advanced rapidly, with parameter scales growing exponentially to reach the trillion-parameter regime ~\cite{brown2020language,PALM,Hoffmann2022}. Training such models requires hundreds to thousands of GPUs, making model parallelism(MP) essential for efficiently distributing computational workloads across devices, typically through tensor parallelism (TP) and pipeline parallelism (PP) ~\cite{narayanan2021,Korthikanti2022}. As model sizes continue to increase, communication overhead, particularly AllReduce operations in TP, has increasingly become a critical performance bottleneck in distributed LLM training ~\cite{ZeRO2020,narayanan2021,shoeybi2020megatronlm,Hosoki2026}.

To mitigate communication bottlenecks, a variety of optimization approaches have been proposed and actively explored in recent years. For DP, gradient compression techniques, including quantization~\cite{SDP4Bit,QSGD,signSGD,TernGrad}, sparsification~\cite{Sparse2017,wangni2018}, and local accumulation~\cite{stich2019localsgd}, have been extensively studied to reduce communication costs. For TP, frameworks such as TACO~\cite{liu2026taco} demonstrate that adaptive FP8-based compression can significantly reduce communication overhead while maintaining near-lossless training accuracy. More broadly, FP8-based methods (e.g., FP8-LM~\cite{peng2023fp8lm} and COAT~\cite{xi2025coat}) improve training efficiency by extending low-precision representations to gradients, optimizer states, and activations. 

However, the limited compression ratio of FP8 is often insufficient to fully alleviate communication bottlenecks at scale, motivating more aggressive low-bit approaches in distributed settings. While 4-bit quantization offers the potential for substantially higher compression, it introduces fundamental challenges beyond simple precision scaling~\cite{llmint82022,frantar2022gptq}. Recent efforts have explored FP4 techniques for model computation~\cite{cao2025metis,castro2026quartet,abecassis2025pretraining} and low-bit communication in distributed training~\cite{SDP4Bit,he2026tahquant}. Yet, these methods primarily target computation, DP or PP, leaving TP communication under extreme low-bit constraints largely unexplored.

Despite its potential to reduce communication overhead, directly applying coarse-grained 4-bit quantization to model-parallel communication faces three key challenges. First, FP4's narrow dynamic range causes long-tail values to overflow and near-zero values to round to zero, distorting gradient directions and destabilizing early-stage training~\cite{Wu2023,liu2026taco}. Second, sub-byte storage requires packing two 4-bit values per byte, while naive packing incurs additional data movement and bit-manipulation overhead, limiting communication efficiency~\cite{Kamath2026}. Third, reducing 4-bit quantization error requires fine-grained quantization, but finer granularity increases scaling-factor computation, potentially making quantization itself a system bottleneck~\cite{zhang2023dual}.
To address these challenges, we propose \textbf{FP4Quant}, a micro-grouped FP4 compression framework that jointly optimizes representation fidelity and GPU execution efficiency for scalable LLM distributed training. The key contributions are summarized as follows:

(1) We propose a \textbf{micro-grouped dual-scale FP4 quantization} mechanism to preserve local tensor distributions under ultra-low-bit compression. Micro-grouping captures local variations, while dual-scale adaptive scaling and max-absolute scaling better match the FP4 dynamic range and reduce overflow.

(2) We introduce a \textbf{hybrid two-level FWHT pipeline} to support fine-grained FP4 compression with limited computational overhead. Intra-warp butterfly operations are efficiently implemented using register-level shuffles, while cross-warp transformations are executed using shared memory, reducing transformation overhead and enabling efficient execution on modern GPUs.

(3) We further develop a \textbf{warp-cooperative FP4 packing} scheme for efficient sub-byte FP4 storage. The kernel exploits warp-level primitives to exchange quantized values and performs nibble-level packing directly in registers, reducing intermediate memory operations and enabling aligned byte-wise storage.

(4) Experiments on NVIDIA H100 clusters show that FP4Quant improves end-to-end training throughput by up to \textbf{1.56×} compared to BF16, while maintaining comparable convergence performance on Qwen models. FP4Quant provides an efficient and scalable solution for ultra-low-bit communication in model-parallel LLM training under practical TP and PP configurations.

\begin{figure}[t]
    \centering
    \includegraphics[width=0.9\linewidth]{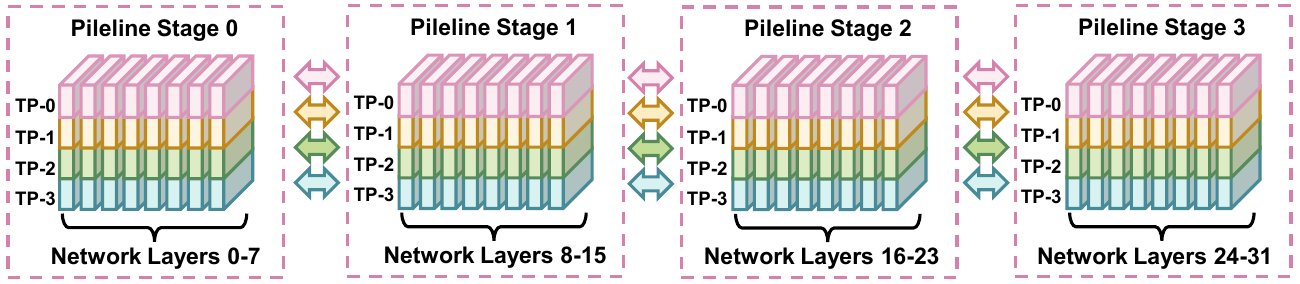}
    \caption{Example distributed training configuration with model parallelism, with 4 tensor-parallel groups, and 2 pipeline stages.}
    \label{fig:MP}
\end{figure}

\section{Background and Preliminaries}
\label{sec:background}

\subsection{Model-Parallel LLM Training}

Model parallelism, including TP and PP, is widely adopted in large-scale LLM training to distribute computation and model parameters across GPUs. 
As shown in Fig.~\ref{fig:MP}, TP partitions individual layers across devices and relies on frequent collective communication operations (e.g., AllReduce) to exchange intermediate tensors~\cite{shoeybi2020megatronlm}, leading to substantial overhead as model size and the degree of parallelism increase~\cite{korthikanti2022reducing}. 
In contrast, PP divides the model into stages and transfers activations and gradients between them, where communication efficiency directly affects pipeline utilization and overall throughput~\cite{huang2019gpipe}. To alleviate these costs, recent studies have explored low-precision compression techniques to reduce communication volume while preserving training accuracy~\cite{peng2023fp8lm,liu2026taco}. However, extending these techniques to ultra-low-bit formats (e.g., FP4) remains challenging, as it is difficult to balance high compression ratios with stable training convergence.

\subsection{Low-Precision Compression for LLM Training}

Recent studies have explored 4-bit floating-point quantization to improve the efficiency of large-scale LLM training. METIS~\cite{cao2025metis} reduces quantization error via spectral-domain FP4 design for accurate W4A4G4 training, while Quartet~\cite{castro2026quartet} enables native FP4 training through hardware-aware optimization. NVIDIA’s NVFP4 pre-training work~\cite{abecassis2025pretraining} demonstrates scalable FP4 training with optimized scaling and mixed-precision scheduling. Subsequent efforts further enhance stability by addressing FP4’s limited dynamic range, including differentiable quantization and outlier-aware scaling~\cite{wang2025optimizing}, as well as adaptive mixed-precision strategies across model components and training stages~\cite{zhou2025pretrainingfp4}.

Beyond computation-oriented FP4 training, several studies have explored low-bit communication compression techniques for distributed LLM training. SDP4Bit~\cite{SDP4Bit} investigates 4-bit gradient communication in sharded data-parallel training and shows that substantial communication reduction can be achieved with limited accuracy degradation. Tahquant~\cite{he2026tahquant} explores low-bit activation compression for pipeline-parallel training to reduce inter-stage communication overhead.
However, existing low-precision techniques primarily focus on computation acceleration, DP or PP communication, whereas the application of ultra-low-bit quantization to TP communication remains relatively underexplored.

\subsection{NVFP4 Representation and Scaling}

NVFP4 is a 4-bit floating-point format designed for efficient AI workloads, retaining sign, exponent, and mantissa components. To compensate for its limited dynamic range, NVFP4 employs block-level scaling, in which multiple FP4 elements share a higher-precision scaling factoGiven an input tensor $X$, NVFP4 represents it as $X \approx s \cdot X_{FP4}$, where $X_{FP4}$ denotes the quantized FP4 values and $s$ is the corresponding scaling factor.
NVFP4 defines two formats, E2M1 and E3M0~\cite{abecassis2025pretraining}, both following the representation $x = (-1)^s 2^{e-B}(1+f)$. E2M1 uses one sign bit, two exponent bits, and one mantissa bit, whereas E3M0 allocates one sign bit and three exponent bits without a mantissa. Despite this design, the extremely limited representation results in rapidly increasing quantization step sizes, leading to larger errors compared with higher-precision formats.

Although block-level scaling improves accuracy over naive FP4 quantization, fixed-size blocks remain insufficient for capturing fine-grained variations in LLM tensors. Large blocks are more sensitive to outliers, whereas smaller blocks incur higher scaling overhead. This trade-off motivates the development of adaptive scaling strategies and efficient GPU implementations to enable accurate and practical NVFP4-based communication.

\begin{figure}[t]
    \centering
    \includegraphics[width=1\linewidth]{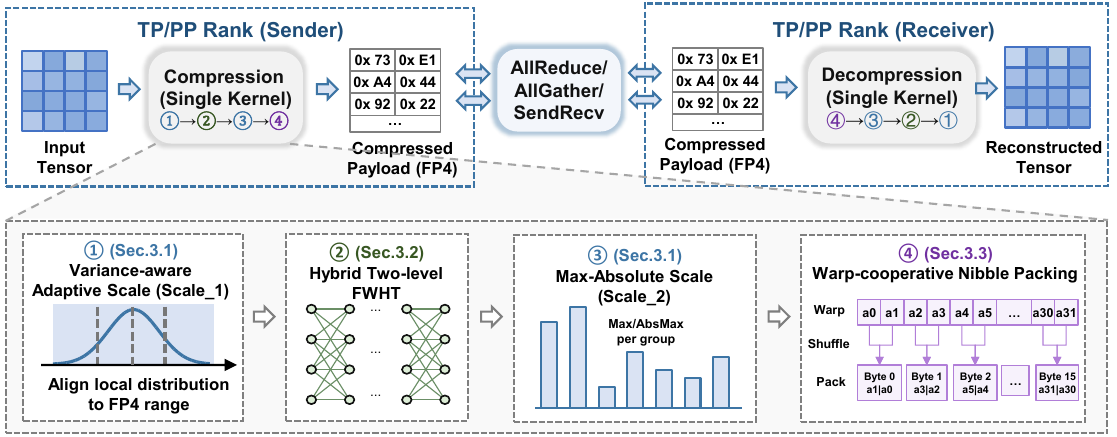}
    \caption{Overview of FP4Quant.}
    \label{fig:overview}
\end{figure}

\section{FP4Quant: Ultra-Low-Precision Communication Framework}
\label{sec:method}

Motivated by the challenges of ultra-low-bit communication, we propose FP4Quant, a micro-grouped FP4 compression framework integrated into the collective communication path. As illustrated in Fig.~\ref{fig:overview}, the compression workflow consists of three main components. Section~\ref{sec:dualscale} describes the micro-grouped dual-scale FP4 quantization mechanism (\ding{172}); Section~\ref{sec:fwht} presents the hybrid two-level FWHT pipeline (\ding{173}); and Section~\ref{sec:packing} introduces warp-cooperative FP4 packing (\ding{174}). At the receiver, the compressed payload is reconstructed through unpacking, dequantization, and the inverse FWHT. Together, these components address the numerical and execution challenges of FP4-based communication compression for model-parallel LLM training.

\begin{figure}[t]
    \centering
    \includegraphics[width=1\linewidth]{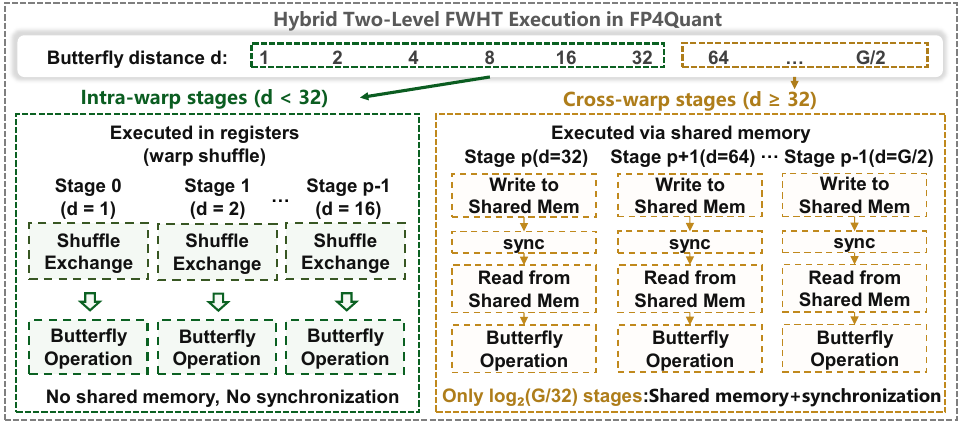}
    \caption{Overview of the hybrid two-level FWHT pipeline.}
    \label{fig:fwht_overview}
\end{figure}

\subsection{Micro-Grouped Dual-Scale FP4 Quantization}
\label{sec:dualscale}

FP4Quant adopts a micro-grouped dual-scale scheme to address the limited dynamic range of NVFP4, combining fine-grained grouping with two-stage scaling for accurate distribution adaptation and efficient FP4 encoding.

\paragraph{Micro-grouping.}
Each communication tensor is partitioned into contiguous micro-groups of $G$ elements, where $G\in{32,64,128,256,512}$. Each group is processed independently to capture localized statistics, enabling fine-grained and adaptive scaling while preserving high-efficiency GPU execution.

\paragraph{Dual-scale quantization.}
For each group $k$, FP4Quant first computes the second moment $\sigma_k^2=\frac{1}{G}\sum_i x_{k,i}^{2}$ and derives an adaptive scale
\begin{equation}
\alpha_k=\mathrm{clip}\left(
\frac{R}{\sqrt{\sigma_k^2+\lambda}},10^{-3},10^{3}\right),
\qquad
\tilde{x}_{k,i}=\alpha_kx_{k,i},
\label{eq:adascale}
\end{equation}
where $\lambda=10^{-6}$ avoids numerical instability. By normalizing the RMS magnitude of each group, $\alpha_k$ maps diverse local distributions into the effective FP4 range. After the FWHT transformation in Section~\ref{sec:fwht}, the transformed values $h_{k,i}$ are quantized using a group-wise max-absolute scale:
\begin{equation}
s_k=\mathrm{clip}\left(\frac{\max_i|h_{k,i}|}{R},10^{-12},10^{6}\right),
\qquad
q_{k,i}=\mathrm{CvtToFP4}
\left(\mathrm{clip}(\frac{h_{k,i}}{s_k},-R,R)\right).
\end{equation}
Here, $\alpha_k$ performs pre-rotation normalization, while $s_k$ determines the final quantization range. Both scales are stored per group and used for reconstruction.

\subsection{Hybrid Two-Level FWHT Pipeline}
\label{sec:fwht}

\paragraph{Motivation.}
FP4 quantization is sensitive to outliers, where a few extreme values can dominate the max-absolute scale and reduce effective precision. FP4Quant applies an orthonormal Walsh--Hadamard transform (FWHT) to redistribute energy across dimensions, producing a more balanced representation (see Fig.~\ref{fig:fwht_overview}). The transform is self-inverse, allowing the same operations for decompression.

\paragraph{Two-level execution.}
A length-$G$ FWHT consists of $\log_2G$ butterfly stages. According to the GPU execution hierarchy, FP4Quant divides these operations into two levels. For intra-warp stages with butterfly distances $d<32$, paired elements are exchanged through register shuffle operations without shared-memory level:

\begin{figure}[t]
    \centering
    \includegraphics[width=0.9\linewidth]{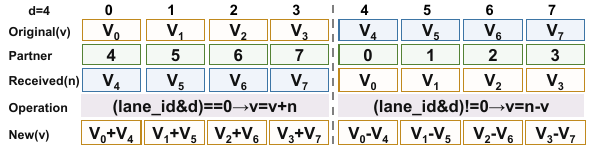}
    \caption{Illustration of the butterfly operation in FWHT.}
    \label{fig:fwht_butterfly}
\end{figure}

\begin{equation}
n \gets \texttt{\_\_shfl\_xor\_sync}(\texttt{0xffffffff},v,d),\qquad
v \gets
\begin{cases}
n-v & (\mathrm{tid}\ \&\ d)\neq0,\\
v+n & \mathrm{otherwise}.
\end{cases}
\label{eq:l1butterfly}
\end{equation}
which corresponds to the butterfly operation shown in Fig.~\ref{fig:fwht_butterfly}. 
These stages are fully unrolled and executed in registers. 
For cross-warp stages with $d\geq32$, intermediate results are exchanged through shared memory with synchronization barriers. Only $\log_2(G/32)$ such stages are required, while groups with $G\leq32$ complete the entire transform using register-level operations. 
The final output is normalized by $1/\sqrt{G}$ to obtain the orthonormal transform.. 
This design aligns FWHT stages with GPU communication primitives, minimizing synchronization overhead while effectively smoothing outliers for FP4 quantization.

\subsection{Warp-Cooperative FP4 Packing}
\label{sec:packing}

FP4 codes occupy four bits, requiring two adjacent codes to be stored in one byte. A straightforward implementation may introduce intermediate memory accesses for byte assembly, where additional loads, shifts, and masks increase memory traffic and reduce the benefit of FP4 compression. FP4Quant therefore performs nibble packing directly in registers using warp-level shuffle operations.

After quantization, thread $i$ holds one 4-bit code $q_{k,i}$ in its low nibble. Adjacent threads exchange their codes through a stride-1 butterfly shuffle:
\begin{equation}
\texttt{hi}_i=
\texttt{\_\_shfl\_xor\_sync}(\texttt{0xffffffff},q_{k,i},1)
\,\&\,\texttt{0x0F}.
\end{equation}
For even lanes ($i\bmod2=0$), the shuffled value provides the high nibble from the adjacent thread, while the local code remains as the low nibble. Each even lane then assembles one byte as $b=(\texttt{hi}_i\ll4)\,|\,q_{k,i}$ and performs a single coalesced store, whereas odd lanes remain inactive. This design avoids shared-memory usage and additional memory round trips, producing naturally aligned byte-level writes across the warp. During decompression, the process is reversed: each thread loads the corresponding packed byte and extracts either the low or high nibble according to the parity of its index before dequantization. 
This warp-cooperative design eliminates intermediate memory traffic and shared-memory usage, achieving near-zero packing overhead with efficient, aligned writes.

\begin{figure}[t]
    \centering
    \includegraphics[width=1\linewidth]{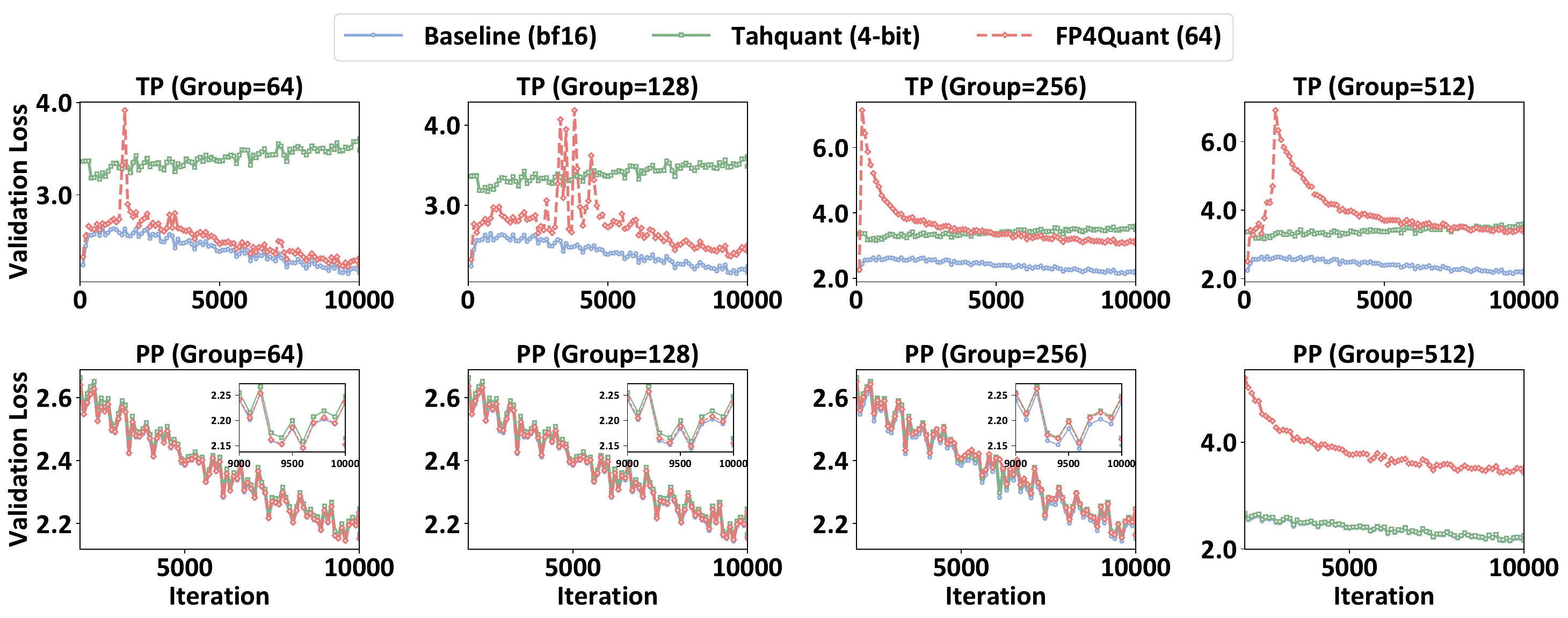}
    \caption{Validation loss of Qwen-7B with different micro-group sizes under TP and PP communication compression.}
    \label{fig:loss_curves}
\end{figure}
% =============================================================================

\section{Evaluation}
\label{sec:eval}

\subsection{Experimental Setup}
\label{sec:setup}

\paragraph{Platform.}
All experiments are conducted on a single node equipped with eight NVIDIA H100 GPUs (80,GB HBM3 each; driver 595.71.05; CUDA 13.2), interconnected via NVLink. FP4Quant is implemented as an NCCL compressor plugin and evaluated within a production-grade model-parallel training stack.

\paragraph{Model, data, and task.}
We train \textbf{Qwen-7B} on the \textbf{OpenWebMath} corpus. Model quality is evaluated using validation loss and the corresponding perplexity (PPL) ($\mathrm{PPL} = e^{\text{loss}}$). System efficiency is measured by end-to-end training throughput (TFLOP/s per GPU) under different parallel settings. Downstream model capability is further evaluated using accuracy on \textbf{CMMLU} and \textbf{C-Eval}.

\paragraph{Parallel configurations.}
We evaluate FP4Quant under three parallel regimes: TP, PP, and their combined MP configuration in large-scale distributed training. For each regime, we investigate micro-group sizes $G \in {64,128,256,512}$ and compare against two baselines: the uncompressed \textbf{BF16 (Ring)} and \textbf{Tahquant}, a representative method for activation communication compression in PP settings.

\begin{table}[t]
  \centering
  \small
  \caption{Downstream accuracy on CMMLU and C-Eval (valid) with $G=64$ under different model-parallel configurations.}
  \label{tab:downstream}
  \setlength{\tabcolsep}{5pt}
  \begin{tabular}{rlcc}
    \toprule
    \textbf{Configuration} & \textbf{Method} 
    & \textbf{CMMLU} & \textbf{C-Eval (valid)} \\
    \midrule
    \multicolumn{2}{c}{BF16 (Ring)}
    & $0.5365\pm0.0045$ 
    & $0.5215\pm0.0133$ \\
    \midrule

    \multirow{2}{*}{TP}
    & FP4Quant 
    & $0.5287\pm0.0045$ 
    & $0.4829\pm0.0133$ \\
    & Tahquant 
    & $0.2530\pm0.0040$ 
    & $0.2771\pm0.0122$ \\
    \midrule

    \multirow{2}{*}{PP}
    & FP4Quant 
    & $0.5288\pm0.0045$ 
    & $0.5342\pm0.0133$ \\
    & Tahquant 
    & $0.5179\pm0.0045$ 
    & $0.5186\pm0.0133$ \\
    \midrule

    \multirow{2}{*}{MP}
    & FP4Quant 
    & $0.5301\pm0.0045$ 
    & $0.5253\pm0.0133$ \\
    & Tahquant 
    & $0.2534\pm0.0040$ 
    & $0.2422\pm0.0117$ \\
    
    \bottomrule
  \end{tabular}
\end{table}

\subsection{Convergence and Model Quality}
\label{sec:eval-quality}

To evaluate the convergence behavior and model quality of FP4Quant under ultra-low-bit communication compression, we measure the validation loss and perplexity across different micro-group sizes ($G\in\{64,128,256,512\}$). Table~\ref{tab:combined_results} summarizes the final validation metrics, while Figure~\ref{fig:loss_curves} presents the convergence curves over $10,000$ iterations under TP and PP configurations.

The results demonstrate that FP4Quant consistently outperforms Tahquant in preserving model quality under FP4 communication compression. Under TP, Tahquant suffers from severe accuracy degradation (loss $3.4848$, $\mathrm{PPL}=32.6163$), whereas FP4Quant with $G=64$ achieves substantially lower loss ($2.2306$, $\mathrm{PPL}=9.3050$) and maintains a stable convergence trajectory close to the full-precision baseline, as illustrated in Fig.~\ref{fig:loss_curves}. Under PP, FP4Quant further demonstrates near-lossless performance (loss $2.1525$, $\mathrm{PPL}=8.6105$), matching BF16 accuracy and outperforming Tahquant.

In addition, FP4Quant exhibits clear sensitivity to group size, where increasing $G$ leads to gradual degradation in TP and a sharp breakdown at large $G$ (e.g., $G=512$) in both TP and PP. This trend suggests that overly coarse grouping introduces scale mismatch and amplifies quantization error. Nevertheless, with appropriately chosen group sizes (e.g., $G\le256$), FP4Quant achieves an excellent balance between accuracy and communication efficiency and consistently delivers superior performance compared to Tahquant.

\subsection{Downstream Task Accuracy}
\label{sec:eval-downstream}

Training loss and downstream accuracy reflect different aspects of model performance: loss measures optimization convergence, whereas downstream tasks evaluate practical model capability. Although the two metrics are not strictly correlated, poor convergence or substantially higher loss than the baseline typically leads to degraded downstream performance. Table~\ref{tab:downstream} reports evaluation results on CMMLU and C-Eval (valid) with $G = 64$. FP4Quant consistently preserves downstream performance across model-parallel settings, achieving scores of $0.5287/0.4829$ (TP), $0.5288/0.5342$ (PP), and $0.5301/0.5253$ (MP), compared with the BF16 baseline ($0.5365/0.5215$). Notably, under PP and MP configurations, FP4Quant matches or slightly exceeds BF16 on C-Eval, demonstrating strong capability preservation under ultra-low-bit compression.

In contrast, TahQuant exhibits severe degradation under communication-intensive settings. Under TP and MP, its performance drops to near random-chance levels ($0.2530/0.2771$ and $0.2534/0.2422$), while remaining stable only under PP. This suggests that coarse-grained 4-bit quantization is insufficient to maintain optimization quality, ultimately degrading downstream performance. Overall, FP4Quant achieves both consistently stable convergence and superior downstream fidelity compared with existing 4-bit communication methods.

\begin{table}[t]
\centering
\small
\setlength{\tabcolsep}{3pt}

\caption{Comparison of training accuracy and throughput between FP4Quant and baselines (BF16 and Tahquant) under TP4 and PP on Qwen-7B.}
\label{tab:combined_results}

\resizebox{\linewidth}{!}{
\begin{tabular}{lcccccccc}
\toprule
\multirow{2}{*}{\textbf{Method}}
& \multicolumn{4}{c}{\textbf{TP}}
& \multicolumn{4}{c}{\textbf{PP}}\\
\cmidrule(lr){2-5}
\cmidrule(lr){6-9}
& \textbf{Loss}$\downarrow$
& \textbf{PPL}$\downarrow$
& \textbf{TFLOP/s}$\uparrow$
& \textbf{Speedup}$\uparrow$
& \textbf{Loss}$\downarrow$
& \textbf{PPL}$\downarrow$
& \textbf{TFLOP/s}$\uparrow$
& \textbf{Speedup}$\uparrow$\\
\midrule

BF16 (Ring)
& 2.1505 & 8.5893
& 111.0 & 1.00$\times$
& 2.1505 & 8.5893
& 189.1 & 1.00$\times$ \\

Tahquant
& 3.4848 & 32.6163
& \textbf{197.4} & \textbf{1.78$\times$}
& 2.1653 & 8.7174
& 189.6 & 1.00$\times$ \\

\midrule

FP4Quant($G=64$)
& \textbf{2.2306} & \textbf{9.3050}
& 155.3 & 1.40$\times$
& \textbf{2.1525} & \textbf{8.6105}
& 190.1 & 1.01$\times$ \\

FP4Quant($G=128$)
& 2.4223 & 11.2713
& 171.2 & 1.54$\times$
& 2.1552 & 8.6298
& \textbf{191.1} & \textbf{1.01$\times$} \\

FP4Quant($G=256$)
& 3.0616 & 21.3613
& \textbf{173.3} & \textbf{1.56$\times$}
& 2.1626 & 8.6940
& 190.7 & 1.01$\times$ \\

FP4Quant($G=512$)
& 3.3512 & 28.5368
& 172.6 & 1.55$\times$
& 3.4270 & 30.7830
& 190.3 & 1.01$\times$ \\

\bottomrule
\end{tabular}
}

\end{table}

\subsection{System Efficiency}
\label{sec:eval-efficiency}

Table~\ref{tab:combined_results} shows that FP4Quant achieves a superior balance between training accuracy and throughput compared to BF16. Under TP, Tahquant attains the highest throughput (197.4 TFLOP/s/GPU, $1.78\times$), but suffers from severe accuracy degradation, indicating that its aggressive quantization compromises training fidelity. 
In contrast, FP4Quant substantially improves model quality while still providing significant speedup. At $G=64$, FP4Quant reduces the loss to 2.2306 while achieving a $1.40\times$ speedup over BF16, and continues to offer a favorable trade-off at larger group sizes. 
For example, at $G=256$, FP4Quant reaches 173.3 TFLOP/s/GPU while achieving significantly better accuracy than Tahquant. Although FP4Quant achieves lower throughput than Tahquant, it avoids severe accuracy degradation while delivering a better accuracy–efficiency trade-off.

Under PP, where computation dominates execution and communication overhead is less significant, all methods achieve comparable throughput (189–191 TFLOP/s/GPU). FP4Quant reaches 190.1--191.1 TFLOP/s/GPU, closely matching the BF16 baseline (189.1 TFLOP/s/GPU) with up to $1.01\times$ speedup, indicating that its quantization and scaling overhead is effectively hidden by computation. Unlike communication-bound TP training, PP offers limited acceleration potential because performance is primarily determined by computation throughput. Overall, FP4Quant enables efficient low-bit communication with substantial acceleration under TP while incurring negligible runtime overhead under PP, demonstrating its effectiveness across diverse parallel configurations.

\section{Conclusion and Future Work}
In this paper, we present FP4Quant, an efficient FP4-based communication compression framework for model-parallel LLM training. It integrates micro-grouped dual-scale quantization, warp-cooperative packing, and a hybrid FWHT-based pipeline to improve quantization fidelity and system efficiency. Experiments show that FP4Quant achieves up to 1.56× end-to-end throughput improvement while preserving convergence.
Our evaluation focuses on Qwen-7B on a single-node eight-GPU H100 system. Future work will investigate scalability across model sizes and multi-node environments, adaptive precision strategies, and applications to heterogeneous systems, multimodal models, and reinforcement learning.

\section*{Acknowledgements}
This work was supported by the National Natural Science Foundation of China through the Excellent Young Scientists Fund Program (Overseas) (Project No. E401470), the National Key Research and Development Program of China (Grant Nos. 2020YFA0907000 and 2025YFB30037002), the National Natural Science Foundation of China (Grant Nos. 62032023 and T2125013), and Ant Group. The AI-driven experiments, simulations, and model training were performed on the robotic AI-Scientist platform of the Chinese Academy of Sciences.

\bibliographystyle{splncs04}
\bibliography{references}

\end{document}